\documentclass{article}

\usepackage[final, nonatbib]{neurips_2021}

\usepackage[utf8]{inputenc} 
\usepackage[T1]{fontenc}    
\usepackage[backref=page]{hyperref}       
\usepackage{url}            
\usepackage{booktabs}       
\usepackage{amsfonts}       
\usepackage{nicefrac}       
\usepackage{microtype}      
\usepackage{epsfig}
\usepackage{graphicx}
\usepackage{amsmath}
\usepackage{amssymb}
\usepackage{booktabs}
\usepackage[table]{xcolor}
\usepackage{color, colortbl}
\usepackage{caption}
\usepackage{ulem}
\graphicspath{{images/class_distribution/},
{images/xviewmbands/}}

\def\imw#1{
\includegraphics[trim=110 45 110 45, clip,width=0.11\textwidth]{images/xviewmbands/#1.png}
}
\def\irow#1{
\imw{#1_rgb}&
\imw{#1_1}&
\imw{#1_2}&
\imw{#1_3}&
\imw{#1_4}&
\imw{#1_5}&
\imw{#1_6}&
\imw{#1_7}&
\imw{#1_8}\\
}

\def\figxviewmbands#1{
\begin{figure*}[#1]
\setlength{\tabcolsep}{1pt}
\begin{tabular}{c|c@{}c@{}c@{}c@{}c@{}c@{}c@{}c}
\irow{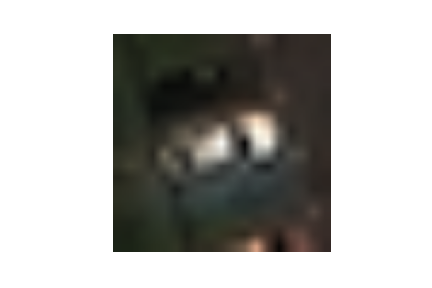}
\irow{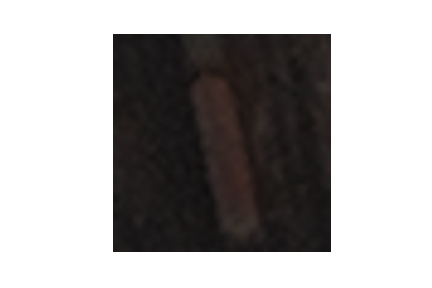}
\irow{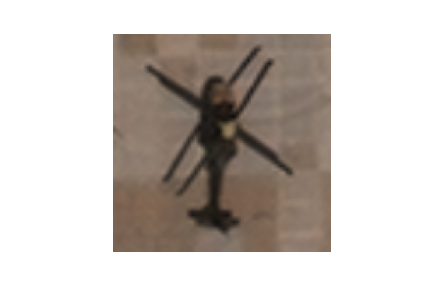}
color RGB &
1.coastal blue&
2.blue&
3.green&
4.yellow&
5.red&
6.red edge&
7.near-IR1&
8.near-IR2\\
\end{tabular}

\caption{An MSI image contains highly correlated channels which collectively reveal structures more clearly than 3-band color images.  {\bf Column 1}: Standard color images of RGB bands.  {\bf Columns 2-9}: Individual channels of increasing wavelengths of an 8-band image, with bands 2,3,5 corresponding to blue, green, and red channels respectively.  Rows 1-3 contain sample instances from head ({\it PV/Personal Vehicle}), body ({\it RV/Rail Vehicle}), and tail ({\it Helicopter}) classes of a highly imbalanced distribution shown in Fig.2a.   A MSI image provides far more information than an RGB image.  For example, in Row 2, the scene structure is hardly visible in the RGB image (Column 1), but fine details are revealed in short-wave bands (Columns 2-5) and the overall shape revealed in long-wave bands (Columns 6-9).
\label{fig:xviewmbands}
\vspace{-4mm}
}
\end{figure*}
}

\def\figxviewTeaser#1{
\begin{figure*}[#1]
\setlength{\tabcolsep}{8pt}
\begin{tabular}{@{}ccc@{}}
\hspace{-0.1in}
\includegraphics[height=0.12\textheight]{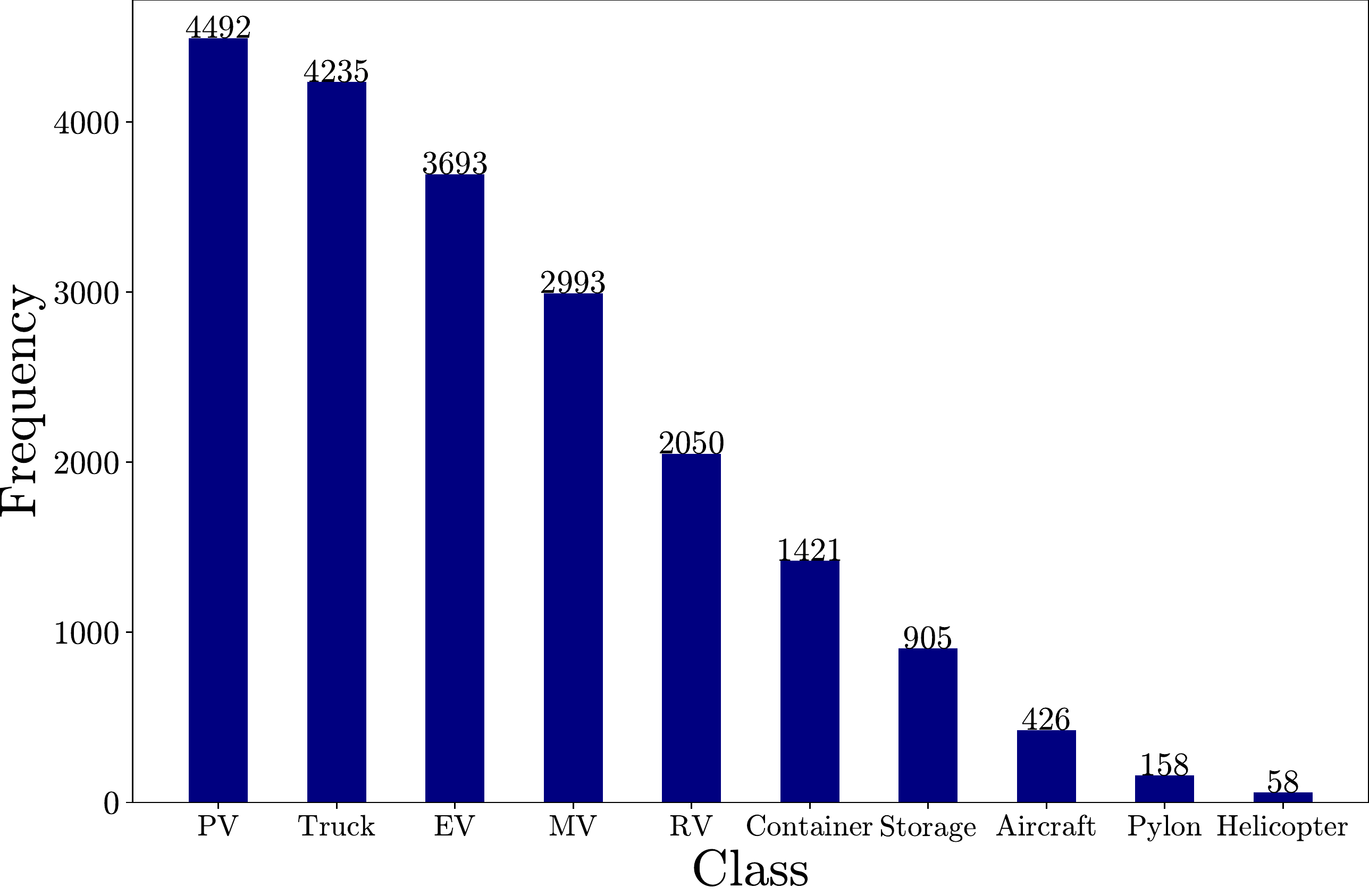}&
\includegraphics[height=0.12\textheight]{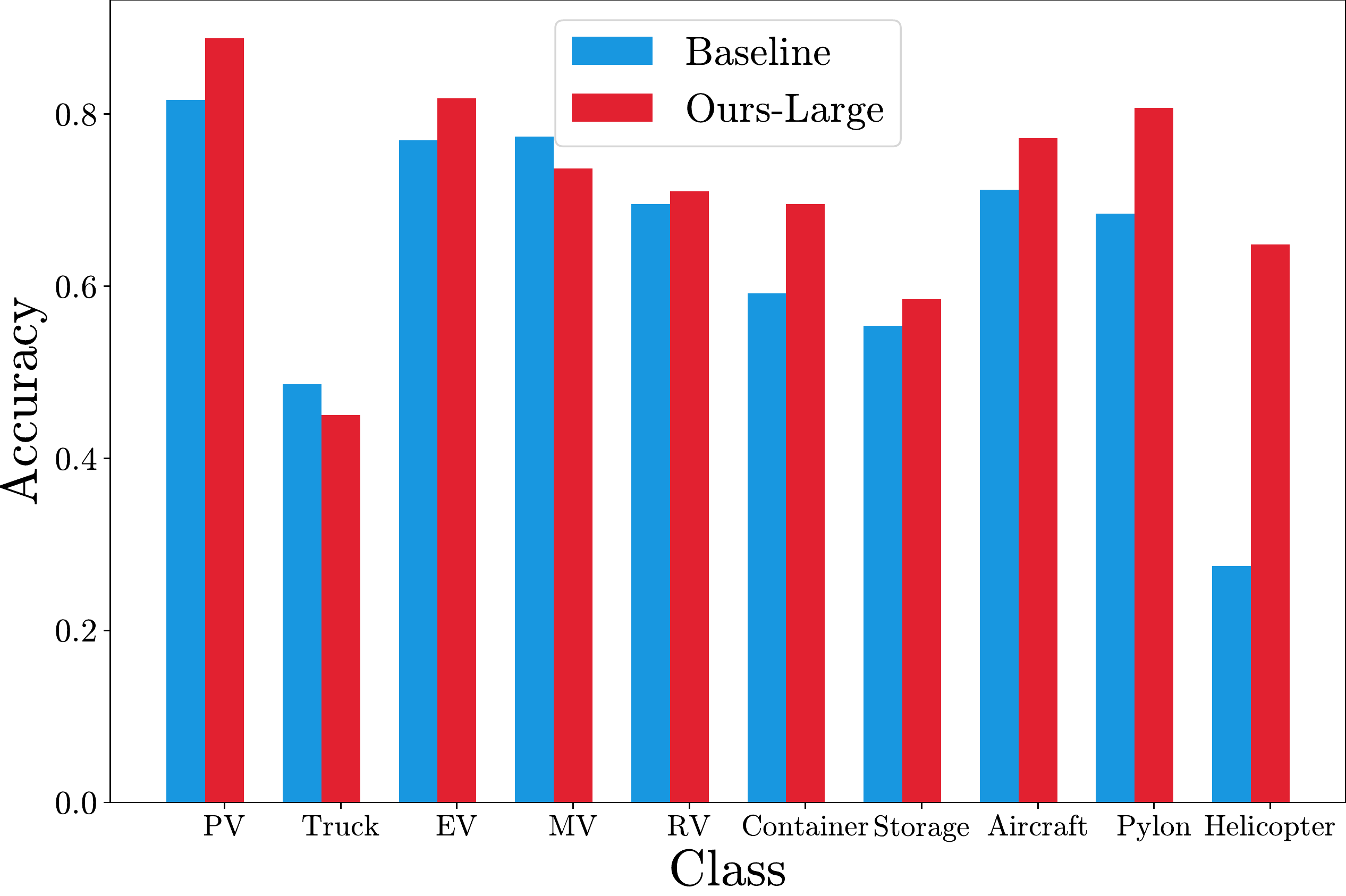}&
\includegraphics[height=0.12\textheight]{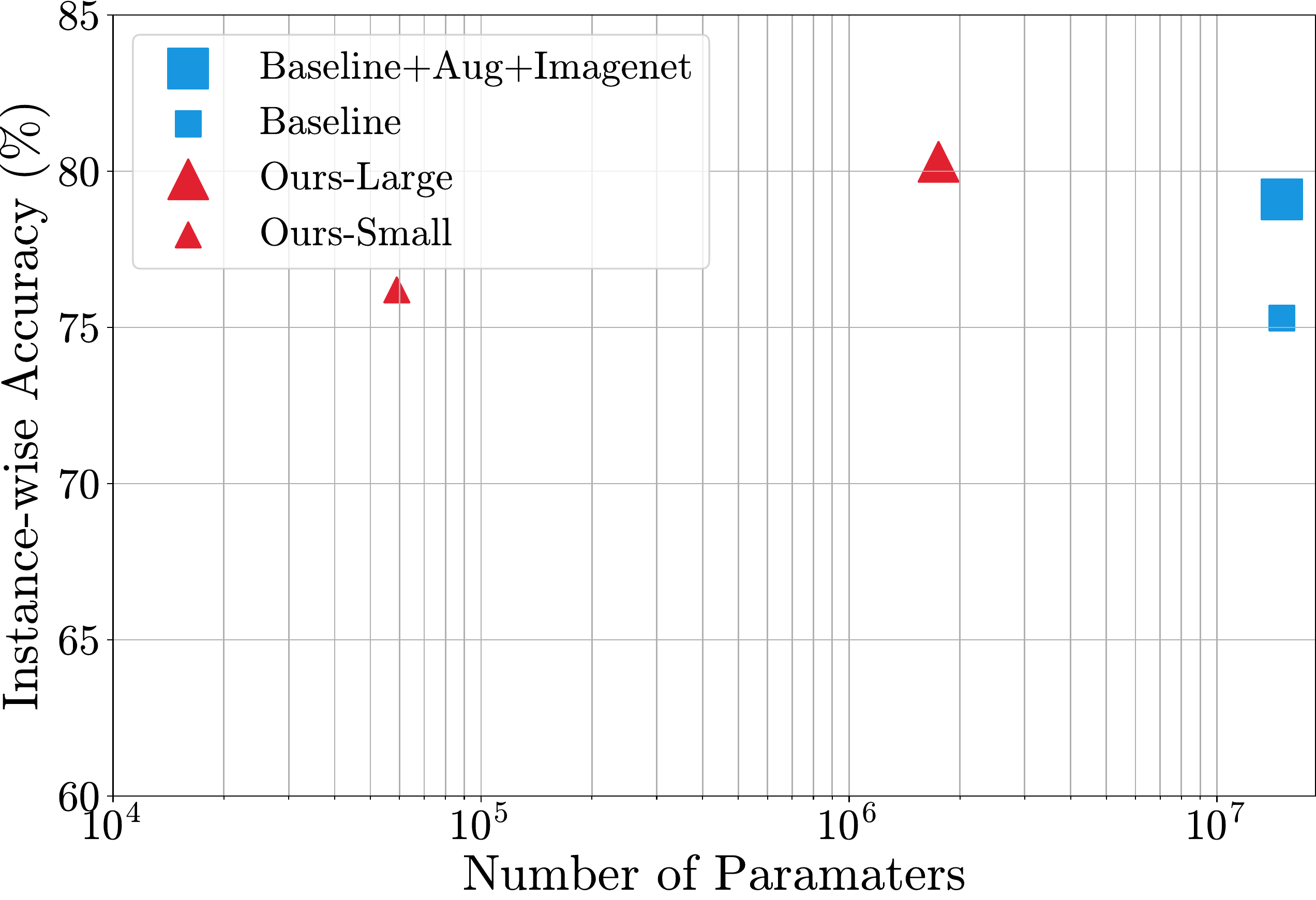}\\
{\bf a)} data distribution &
{\bf b)} classwise accuracy &
{\bf c)} model size vs. overall accuracy\\ 
\end{tabular}

\caption{
We apply complex-valued learning method CDS to real-valued image classification of xView MSI data.  
{\bf a)} The data is highly imbalanced over 10 classes.
{\bf b)} Bar graph of class-wise accuracy. Our method outperforms the ResNet baseline over 8 out of 10 categories, with the largest difference coming from the hardest category with the fewest examples.
{\bf c)} Scatter plot showing parameters and instance-wise accuracy. Our models are ultra-lean compared to the ResNet baseline and deliver comparable accuracy without using augmentation or pretraining. Ours-Small uses only $0.5\%$ of the number of parameters of the baseline, yet outperforms the naive baseline and achieves $95\%$ accuracy as the baseline which requires both ImageNet pretraining and data augmentations.
\vspace{-5mm}
\label{fig:xviewTeaser}
}
\end{figure*}
}

\title{Multi-Spectral Image Classification with \\
Ultra-Lean Complex-Valued Models}

\author{%
\setlength{\tabcolsep}{7pt}%
\begin{tabular}{@{}ccccc@{}}
Utkarsh Singhal$^1$&
Stella X. Yu$^{1,2}$& 
Zackery Steck$^3$&
Scott Kangas$^3$&
Aaron A. Reite$^4$\\[3pt]
\end{tabular}
\\
\setlength{\tabcolsep}{8mm}
\begin{tabular}{@{}cc@{}}
$^1$ UC Berkeley / ICSI &
$^2$ University of Michigan  \\
$^3$ Etegent Technologies &
$^4$ National Geospatial-Intelligence Agency\\
\end{tabular}
}

\begin{document}
\maketitle
\vspace{-3mm}
\begin{abstract}
Multi-spectral imagery is invaluable for remote sensing due to different spectral signatures exhibited by materials that often appear identical in greyscale and RGB imagery. Paired with modern deep learning methods, this modality has great potential utility in a variety of remote sensing applications, such as humanitarian assistance and disaster recovery efforts. State-of-the-art deep learning methods have greatly benefited from large-scale annotations like in ImageNet, but existing MSI image datasets lack annotations at a similar scale. As an alternative to transfer learning on such data with few annotations, we apply complex-valued co-domain symmetric models to classify real-valued MSI images. Our experiments on 8-band xView data show that our ultra-lean model trained on xView from scratch {\it without} data augmentations can outperform ResNet {\it with} data augmentation and modified transfer learning on xView. Our work is the first to demonstrate the value of complex-valued deep learning on real-valued MSI data.
\end{abstract}

\vspace{-3mm}
\section{Application Context}
\vspace{-2mm}
Multispectral (MSI) remote sensing images play a crucial role in environmental monitoring and humanitarian assistance and disaster recovery (HADR) applications. These images contain multiple bands in addition to the Red (R), Green (G), and Blue (B) bands of standard color images. Each band captures a unique part of the electromagnetic (EM) spectrum, allowing a properly trained model to finely distinguish between materials that look identical in greyscale or RGB imagery. Combining MSI with the recent progress in computer vision tasks such as classification and detection thus has a huge potential to automate and improve large-scale disaster assessment. However, applying standard computer vision techniques such as supervised learning and transfer learning to MSI is severely limited by the absence of large-scale fully annotated MSI datasets. We demonstrate that better architectures allow for accurate aerial image recognition without pretraining or data augmentation.
\vspace{-2mm}
\section{Introduction}
\vspace{-2mm}
 While MSI data are readily available at a large scale, fully annotated datasets are scarce and lack diversity compared to large scale RGB datasets such as ImageNet \cite{Shermeyer2021RarePlanesSD, Chiu2020AgricultureVisionAL, Christie2018FunctionalMO}. It has been widely shown that models pretrained on ImageNet \cite{deng2009imagenet} can be effectively fine-tuned with only a small annotated set to work on images from novel domains such as line drawings and medical images, e.g., magnetic resonance imaging (MRI) and Chest X-Ray \cite{2017chexnet}, etc. For domains with one rather than three channels, a standard practice is to duplicate the single channel three times so the image may be input into a standard RGB pretrained model \cite{helber2017}. Unlike these image domains, MSI images have more input channels than the three RGB channels, so ImageNet pretrained models cannot be transferred directly.

There are a few applicable approaches to handle this problem: One could apply popular unsupervised representation learning methods \cite{ufl, he2019momentum, simclr, byol, wang_cld} to unlabelled MSI data to produce the trained model. However, training from scratch on a large-scale dataset is computationally expensive, often requiring many end-to-end training cycles to tune hyperparameters before producing a pretrained model on par with ImageNet-supervised pretraining. Alternatively, the number of channels may be reduced to 3 so that an ImageNet pretrained model can be utilized. Typically, 3-band combinations are manually selected \cite{xie2015,penatti2015,castelluccio2015}, although other channel reduction techniques may also be used, such as averaging neighboring spectral bands or training a 1x1 convolutional kernel. We explore several such methods as competitive baselines for our proposed approach.
\figxviewmbands{tp}
\figxviewTeaser{tp}

We investigate a drastically different approach based on a recent proposal \cite{cds2022} that represents standard 3-channel color images as two-channel complex-valued images and applies co-domain symmetric (CDS) neural networks. This approach exploits the complex-valued scaling symmetry present in the data, producing leaner models with higher accuracy on RGB images and complex-valued data \cite{cds2022, chakraborty2019surreal, trabelsi2017deep}. Specifically,  given image $I$ of $m$ channels,  we use sliding-channel encoding to turn it into a complex-valued image of  $m\!-\!1$ channels: \begin{align}
I &= [I_1, I_2, \ldots, I_m]
\rightarrow 
[I_1 + i I_2, \, \, \, I_2 + i I_3,  \, \, \,  \ldots, \, \, \,  I_{m-1} + i I_m]
\end{align} where $i$ is the imaginary number unit $\sqrt{-1}$.
Composing two adjacent channels into one complex-valued channel  binds adjacent channels, imparting an ordering absent in the original MSI channels. Additionally, the phase in this representation encodes the relative value of adjacent channels.

On the 8-band MSI images corresponding to the publicly available RGB satellite image dataset xView \cite{xview}, we demonstrate that Co-domain Symmetry (CDS) \cite{cds2022} delivers an ultra-lean neural network trained from scratch that performs better than ImageNet pretrained ResNet model baselines that are much larger and require data augmentation. Our work demonstrates the value of complex-valued CDS models on real-valued MSI images with strong natural correlations across channels.


\vspace{-2mm}
\section{Related Work}
\vspace{-2mm}
\textbf{Complex Valued Deep learning}: Complex-valued deep neural networks have been explored since the early days of deep learning. \cite{nitta2003xor} analyzes the decision boundary of complex-valued neurons, discovering that the real and imaginary components have orthogonal boundaries. This property can be used to gain higher representational capacity along with better regularization. \cite{yu:bright09} use complex numbers to encode confidence and ordering of depth data, producing a significantly more robust criterion for integrating noisy local orderings. Deep Complex Networks \cite{trabelsi2017deep} convert modern CNN architectures like ResNet to complex-valued models, examining and fixing the resulting convergence issues in the process.  CDS \cite{cds2022} builds complex-scale equivariant and invariant layers to tackle the problem of complex-valued scaling ambiguity. CDS also introduces new architectures which compose the invariant/equivariant layers to achieve complex-scale invariance. These properties allow for higher accuracy with significantly leaner models. Our work uses the CDS-Large model.

\textbf{Deep Learning for MSI}: 
MSI has become more prevalent with the launch of the European Space Agencies Sentinel-2A and 2B satellites, as well as Maxar's Worldview-3 \cite{Shermeyer2021RarePlanesSD, Chiu2020AgricultureVisionAL, Christie2018FunctionalMO}.  However, computer vision on MSI data and applications remains under-explored compared to RGB imagery and large-scale, diverse annotated datasets comparable to ImageNet are lacking. \textit{Agriculture vision} \cite{Chiu2020AgricultureVisionAL} contains $94,986$ images of farmlands with semantic segmentation annotations, but only contains 4 channel (RGB-IR) images. \textit{Functional map of the world (fMOW)} contains over 1 million annotated image \textit{pairs}, but a significant portion of these are 3/4-band only, and the task is multi-view classification. RarePlanes \cite{Shermeyer2021RarePlanesSD} contains a mixture of real and synthetic images in 1,3 and 8 bands, but lacks diversity as it focuses on subcategories of only one main category -- planes. Recent research has focused on transfer learning approaches applying convolutional neural networks (CNNs) pretrained on ImageNet to MSI overhead imagery. \cite{xie2015,penatti2015,castelluccio2015}.  More recent work fine-tuned the pretrained network by re-initializing the last layers of the network and training them on annotated satellite images \cite{helber2017}.

\vspace{-4mm}
\section{Methods}
\vspace{-2mm}
We use the CDS model proposed in \cite{cds2022}. This work views complex-valued scaling as a co-domain transformation, creating novel equivariant/invariant layer functions and architectures that exploit co-domain symmetry. Specifically, we use the \textit{CDS-Large} architecture with the \textit{Sliding Encoding} to encode the input channels. However, unlike CDS, we apply this network to the imbalanced 8-band classification setting and tackle the class imbalance problem. We summarize the framework and the necessary modifications such as post-hoc logit adjustment that improve performance on class imbalanced datasets \cite{menon2021logadj}. We refer the reader to \cite{cds2022, menon2021logadj} for details about CDS and logit adjustment.

\textbf{Complex-valued scaling}: Complex-valued scaling describes the set of transformations that multiplies the input image with a complex-valued scalar. Given a complex-valued input image $x \in \mathbb{C}^{H \times W \times C} $ and a complex-valued scalar $s \in \mathbb{C}$ the corresponding transformation $T_s$ is defined as $ T(x) = s.x$. This transformation appears in naturally complex-valued data and is equivalent to important transformations in complex-valued representations of real-valued data. For example, hue shifts in color images can be described as complex-valued scaling in the complex-valued LAB encoding \cite{cds2022}.

\textbf{Co-Domain Symmetry}: Co-domain symmetric models use layers that are equivariant/invariant to complex-valued scaling of the input. An invariant layer is defined as a function $f$ such that for any complex-valued input image $x$ and complex-valued scalar $s$, the layer follows the property $f(s. x) = f(x)$. An equivariant layer, on the other hand, follows $f(s . x) = s . f(x)$. By combining these layers into invariant architectures, CDS \cite{cds2022} builds models which exploit the co-domain symmetries present in the data. This allows for leaner models that learn from less data and require no augmentation. We use the CDS-Large model. Please see Table \ref{tab:ourBIG} for architecture details.

\textbf{Sliding Encoding}: The \textit{sliding encoding} proposed in \cite{cds2022} encodes the channels of an input image real-valued image $x \in \mathbb{R}^{H \times W \times C} $ into a complex-valued image $z \in \mathbb{R}^{H \times W \times C-1}$ by pairing adjacent real channels as the real and imaginary parts of a complex channel. Specifically, given channels $[I_1, I_2, I_3, \ldots, I_m]$, the sliding encoding represents it as $[I_1 + iI_2, I_2+iI_3, \ldots, I_{m-1}+iI_m]$. The phase of the resultant complex channel thus represents the intensity ratio between the neighboring real channels.  This representation naturally encodes the intrinsic order of (increasing) spectral frequencies, whereas the standard real MSI representation treats each band independently.

\textbf{Logit Adjustment}: CDS can generalize better with leaner models and no data augmentation in settings like CIFAR-10. However, this method has not previously been explored for the imbalanced classification setting such as the proposed xView 8-band dataset. To adapt the model to this setting, we use post-hoc logit adjustment \cite{menon2021logadj}. We start by measuring the number of examples in each class and normalizing the result to get class the probability vector $[p_1, p_2, \ldots, p_N] \in \mathbb{R}^N$. Then given a vector $[l_1, l_2, \ldots, l_N] \in \mathbb{R}^N$ containing the predicted logits for $N$ classes, post-hoc logit adjustment simply produces the prediction: $[l_1-\tau \log p_1, l_2-\tau\log p_2, \ldots, l_3 - \tau \log p_N]$ 
where $\tau \in \mathbb{R}$ decides the amount of adjustment. We set $\tau = 1$ for all our experiments.

\vspace{-4mm}
\section{Experiments}
\vspace{-4mm}
We compare baselines with our proposed method on the task of imbalanced classification on the xView 8-band image dataset. We compare the overall accuracy and classwise accuracy of each model.

\textbf{xView 8-band}: xView is a large-scale 8-band MSI dataset with bounding box annotations for each target. Each channel in an 8-band image consists of measurements in a different band of the electromagnetic spectrum. The additional bands allow for additional discrimination ability (see Figure \ref{fig:xviewmbands}). For instance, the ratio of Near-IR and IR bands can be used to monitor plant health since minute changes in the latter are clearly visible in the former (while invisible in RGB). 

We derive an MSI classification dataset from the 8-band WorldView 3 images corresponding to the bounding box annotations of selected categories. xView consists of $60$ total classes, from which we select $10$ supercategories: StorageTank, Helicopter, Pylon, Maritime Vessel (MV), ShippingContainer, Fixed-Wing Aircraft (FWAircraft), Passenger Vehicle (PV), Truck, Railway Vehicle (RV), and Engineering Vehicle (EV). These categories cover vehicles, buildings, and objects of various sizes. These categories were chosen as they contained a large number of images with our required criteria. To make the class distribution more reasonable, we limit the number of samples per category to 5000. This dataset is highly imbalanced, with the largest classes containing $60\times$ more examples than the smallest classes, thus representing a challenging classification problem.

 We resampled the full scenes to a standard 30 cm GSD to limit variations in the target scale. We further cropped square image chips centered around each bounding box using the box’s longest dimension and padded the box with $\pm 5$ pixels. Finally, we discarded chips that crossed image boundaries and scaled the pixel values using the pixel maximum of $6338$ to ensure each channel ranges from $[0, 1]$. Finally, we resized the images to 32x32. We created the train and validation partitions by performing a 90:10 split of the chips generated from the training dataset, resulting in $20431$ training samples and $2270$ validation samples. We perform no subsampling on the test set, resulting in $63279$ test samples. Please refer to Figure \ref{fig:xviewTeaser} for details. 

\textbf{Baselines:} For baselines we use ResNet18 pretrained on ImageNet (ILSVRC2012) and adapted for transfer learning on 8-band imagery using 3 different methods: \textbf{1x1 Conv}: This variant uses a single learned 1x1 convolution prepended to the ResNet18 backbone that reduces the 8-band input into a 3-band input compatible with ResNet18's 3-channel input. Please refer to Figure \ref{fig:xviewrgbconv} for a visualization of this layer's output after training. \textbf{8-band stem}: This approach replaces the standard 3-channel input layer of the ResNet18 with an 8-channel input layer, convolving with the standard 7x7 kernel with stride=2 and padding=3. \textbf{Deep 8-band stem}: This approach replaces the standard single-layer stem with 3 convolutional layers with size $3$ kernels and stride $2$ input layer. The intermediate layers apply BatchNorm and ReLU. In addition to the modifications to the model architecture, we implemented two data-centric approaches that modify the input 8-channel data to be compatible with a traditional 3-channel network. \textbf{Average}: In this approach, we average the channels together to produce a single-channel image, then duplicate those to produce a 3-channel image. \textbf{Binned average}: An additional suitable approach is leveraging the spectral relationships between the 8-channels themselves- namely their location on the spectrum. We group the channels together into 3 bins and average within each bin to produce a 3-channel image. In our implementation, we binned channels $[1, 2, 3]$ (Coastal Blue, Blue, Green), $[4, 5, 6]$ (Yellow, Red, Red Edge), and $[7, 8]$ (Near-IR1, Near-IR2) together.

\textbf{Proposed Method}: We use the CDS-Large architecture. We scale the images down to $32\times 32$ pixels in size and encode the input with the \textbf{sliding encoding}, converting it into a $7$-channel complex-valued input. We also modify the first convolution layer of CDS-Large to have $7$ complex-valued input channels. We train the model using the AdamW optimizer with batch size $16$, learning rate $10^{-4}$, and weight decay $10^{-2}$ for $15000$ batches. We validate the model after every $200$ steps, choosing the model with the best validation accuracy. Unlike the baselines, this model is trained without data augmentation. To address the imbalance in the training data, we apply post-hoc logit adjustment \cite{menon2021logadj}. For further analysis of model size and accuracy, we also add the \textit{CDS-I} model \cite{cds2022} as an ablation.

\textbf{Results:} We compare each method based on two metrics: \textit{instance-wise accuracy} (I-Acc) and \textit{class-wise accuracy} (C-Acc) to evaluate generalization ability and susceptibility to imbalanced classes respectively. Instance-wise accuracy computes the total proportion of correctly classified instances. However, for imbalanced classification, this metric is not representative of accuracy on tail classes. As a result, we also separately compute the accuracy for each class and report the average.

\begin{table}[tp]
\centering
    \begin{tabular}{@{}c@{}}
    \begin{tabular}{ l | r | r r} 
     \toprule
     \textbf{Our(no pretrain)} & Params & I-Acc & C-Acc \\ [0.5ex] 
     \midrule
     \rowcolor{black!10}
    \multicolumn{4}{c}{\textit{\text{Larger Model (CDS-: \cite{cds2022})}}}\\
     \rowcolor{white}
     CDS-L & 1.75M       & \textbf{80.3} & \underline{63.2} \\ 
     CDS-L + LA & 1.75M  & \underline{79.9} & \textbf{71.1} \\ 
     \midrule
    \rowcolor{black!10}
    \multicolumn{4}{c}{\textit{\text{Smaller Model (CDS-E \cite{cds2022})}}}\\
    \rowcolor{white}
     CDS-E       & 59.4k & 76.2 & 57.3 \\ 
     CDS-E + LA  & 59.4k & 73.4 & 62.6 \\ 
     
     \bottomrule
    \end{tabular} \label{tab:our_results} \\
     \\ 
    \begin{tabular}{ l | r | r r} 
     \toprule
     \textbf{R18 (pretrain)} & Params & I-Acc & C-Acc \\ [0.5ex] 
     \midrule
     \rowcolor{black!10}
    \multicolumn{4}{l}{\textit{\text{Architecture Modification}}}\\
    \rowcolor{white}
     1x1 Conv         & 11.5 M & 77.0 & 72.0 \\ 
     8-band Stem     & 11.5 M & 78.8 & 72.4 \\ 
     Deep 8-band & 11.5 M & 77.1 & 70.4 \\ 
     \midrule
    \rowcolor{black!10}
    \multicolumn{4}{l}{\textit{\text{Data Modification}}}\\
    \rowcolor{white}
     Average           & 11.5 M & 76.3 & 68.0 \\ 
     Binned Avg.   & 11.5 M & \textbf{79.1} & \textbf{72.9} \\
     \bottomrule
    \end{tabular}  \label{tab:pret_results}   \\
     \\
    \end{tabular}
\begin{tabular}{@{}cc@{}}
        \begin{tabular}{ l | r r} 
         \toprule
        \textbf{ R18 (no pretrain)} & I-Acc & C-Acc \\ 
         \midrule
         \rowcolor{black!10}
        \multicolumn{3}{l}{\textit{\text{Architecture Modification}}}\\
        \rowcolor{white}
         1x1 Conv & 75.0 & 60.2 \\ 
         1x1 Conv (LA)  & 73.7 & 65.3 \\ 
         8-band Stem & 75.2 & 60.3\\ 
         8-band Stem (LA) & 74.0 & 63.5 \\ 
         Deep 8-band Stem & \textbf{75.3} & 61.2 \\ 
         Deep 8-band Stem (LA) & 74.2 & \textbf{64.9} \\

         \midrule
        \rowcolor{black!10}
        \multicolumn{3}{l}{\textit{\text{Data Modification}}}\\
         \rowcolor{white}
         Average & 72.3 & 56.2 \\ 
         Average (LA) & 70.0 & 58.0 \\ 
         Binned Average & 74.3 & 60.0 \\

        Binned Average (LA) & 73.1 & 62.6 \\ 
        \bottomrule
        \end{tabular} \label{tab:nopret_results} \\
        \\
    \end{tabular}
    \vspace{-3mm}
    \caption{Accuracy of each model under different training settings. Our ultra-lean models match the baselines without data augmentation or pretraining. (LA = Post-hoc logit adjustment. I-Acc = Instance-wise accuracy, C-Acc = Class-wise accuracy). \textbf{(a)} Our proposed models are trained without any augmentation or pretraining. We use the CDS-Large and CDS-E models from \cite{cds2022}. Our large model is competitive with ResNet18 without augmentation or pretraining and uses only $15\%$ of the parameters. Our smaller model can beat some ResNet-18 baselines on instance-wise accuracy while using $200\text{x}$ fewer parameters. \textbf{(b)} ResNet18 baseline with ImageNet pretraining, augmentations, and logit adjustment. \textbf{(c)} ResNet18 baseline without ImageNet pretraining and augmentations.
\label{Tab:8band}
}
\vspace{-10mm}
\end{table}

\def\imws#1{
\includegraphics[trim=100 45 100 45, clip, width=0.25\textwidth]{images/1x1conv_visualizations/#1.png}
}
\def\irows#1{
\imws{nonpretrained/#1_rgb}&
\imws{nonpretrained/#1_1x1conv}&
\imws{pretrained/#1_1x1conv} \\
}

\def\figxviewrgbvsconv#1{
\begin{figure}[#1]
\centering
\setlength{\tabcolsep}{0pt}
\begin{tabular}{@{}c@{\hspace{10pt}}|@{\hspace{20pt}}cc@{}}
\irows{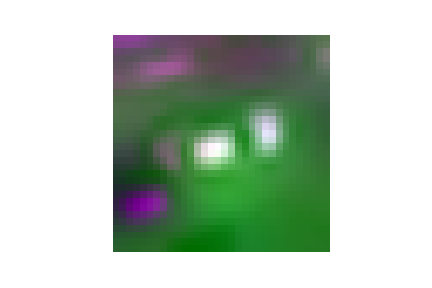}
\irows{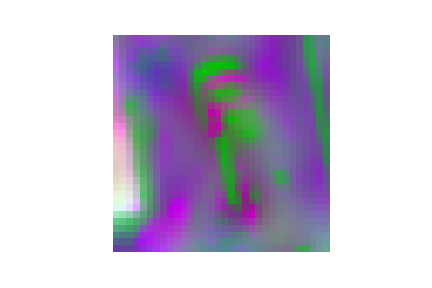}
\irows{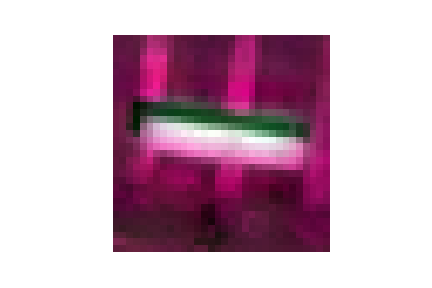}
\irows{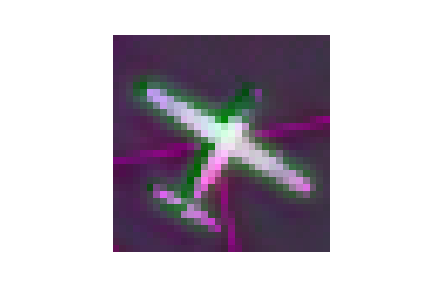}
\irows{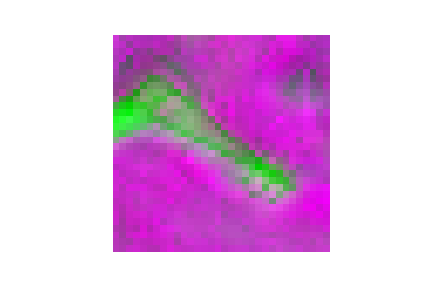}
\irows{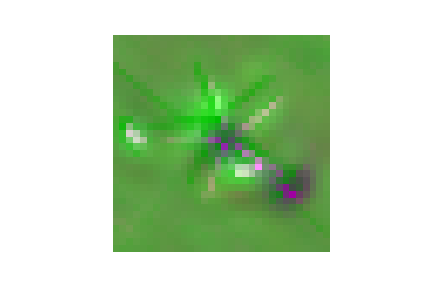}
\textbf{original RGB} & \textbf{8-to-3 bands: initial} & \textbf{8-to-3 bands: final learned} \\
\end{tabular}
\caption{8-channel input contains useful discriminative information. The 1x1 convolutional filter prepended to the Resnet18 is learning to squeeze the 8-channel input into 3 channels. {\bf Column 1}: Standard color images of RGB bands.  {\bf Column 2}: Color visualization of the 3-channel output from the 1x1 convolutional filter for randomly initialized Resnet18. {\bf Column 3}: Color visualization of the 3-channel output from the 1x1 convolutional filter for Imagenet1k pretrained Resnet18. Rows 1-6 contain sample instances from head ({\it PV}, {\it EV}), body ({\it ShippingContainer}) and tail ({\it FWAircraft/Fixed Wing Aircraft}, {\it Pylon}, {\it Helicopter}) classes of the distribution. The final projected 8-band image is able to distinguish between different surfaces and materials where RGB fails.
\label{fig:xviewrgbconv}
}
\end{figure}

}

Table \ref{Tab:8band}  shows each model's instance-wise and class-wise accuracy on the xView dataset. Table \ref{Tab:8band}c shows ResNet18 accuracy with and without logit adjustment for different types of architecture/data modifications, whereas \ref{Tab:8band}b shows Resnet18 accuracy for a model pretrained with ImageNet, trained with Augmentation, and evaluated with Logit Adjustment. \textbf{Instance-wise accuracy}: We find that our model (Table \ref{Tab:8band}), without any augmentation or pretraining and with significantly fewer parameters, can outperform ResNet18. More strikingly, even the ultra-lean model with only $58k$ parameters can outperform the baselines. \textbf{Class-wise accuracy}: Our logit-adjusted large model outperforms the ResNet baselines without ImageNet pretraining and augmentations, and is competitive with the augmented and logit-adjusted baseline while using only $15\%$ of the parameter count.

\textbf{Conclusion:} We apply complex-valued co-domain symmetric models to classify real-valued MSI images. Our xView experiments demonstrate that our lean models trained from scratch without data augmentations or pretraining can outperform ResNet18 with data augmentation and transfer learning. Our small model achieves comparable accuracy while using $200$x fewer parameters, and our large model achieves higher instance-wise and comparable class-wise accuracy. These results indicate the promise of lean models with built-in symmetries applied to novel domains.

\textbf{Acknowledgements:} This research was supported, in part, by the National Geospatial Intelligence Agency / Etegent Technologies Ltd. contract HM047622C0004; approved for public release, NGA-U-2022-02201.

{\small
\bibliographystyle{unsrt}
\bibliography{mband}
}
\clearpage
\newpage

\section{Supplementary}
This section contains (a) visualization of the the 3-channel output of the 1x1 convolutional filter for ResNet18 as well as (b) table describing the exact model architecture used by CDS-Large. The 3-channel visualization demonstrates the utility of MSI input by visualizing the information projected into RGB. Each band captures a unique part of the electromagnetic spectrum, and each material has a unique spectral profile. This results in an overall image with significantly more contrast between different materials than an RGB image.

\figxviewrgbvsconv{}

\clearpage

\begin{table}[h!]\centering
\caption{CDS-Large model architecture, reproduced from \cite{cds2022} with permission.
\label{tab:ourBIG} 
}
\setlength{\tabcolsep}{1pt}
\begin{tabular}{@{}cccccc@{}}
 \toprule
 {\bf Layer Type} & {\bf In Shape} & {\bf Kernel} & {\bf Out Shape}\\
  \midrule
Econv        &\([3, 32,32]\)	& $3$	 &\([64, 32, 32]\)\\
\midrule
Conjugate Layer       &\([64,32,32]\)   & $1$              &\([64,32,32]\)\\
\midrule
Econv (Groups=2)        &\([64, 32,32]\)	& $3$   &\([64, 32, 32]\)\\
\midrule
ComplexBatchNorm        &\([64, 32,32]\)	& -	  &\([64, 32, 32]\)\\
\midrule
{C}ReLU        &\([64, 32,32]\)	& -	&\([64, 32, 32]\)\\
\midrule
Econv (Groups=2)        &\([64, 32,32]\)	& $3$	   &\([128, 32, 32]\)\\
\midrule
ComplexBatchNorm        &\([128, 32,32]\)	& -	   &\([128, 32, 32]\)\\
\midrule
{C}ReLU        &\([128, 32,32]\)	& - &\([128, 32, 32]\)\\
\midrule
Eq. MaxPool     &\([128, 32,32]\)	& $2$	 &\([128, 16, 16]\)\\
\midrule
ResBlock(groups=2)      &\([128,16,16]\)	& -  &\([128, 16, 16]\)\\
\midrule
Econv (Groups=4)        &\([128,16,16]\)	& $3$	  &\([256,16,16]\)\\
\midrule
ComplexBatchNorm        &\([256,16,16]\)	& -	&\([256,16,16]\)\\
\midrule
{C}ReLU        &\([256,16,16]\)	& -   &\([256,16,16]\)\\
\midrule
Eq. MaxPool     &\([256,16,16]\)	& $2$	  &\([256,8,8]\)\\
\midrule
Econv (Groups=2)        &\([256,8,8]\)	& $3$	  &\([512,8,8]\)\\
\midrule
ComplexBatchNorm        &\([512,8,8]\)	& -  &\([512,8,8]\)\\
\midrule
{C}ReLU        &\([512,8,8]\)	& -	   &\([512,8,8]\)\\
\midrule
Eq. MaxPool     &\([512,8,8]\)	& $2$	   &\([512,4,4]\)\\
\midrule
ResBlock(groups=4)      &\([512,4,4]\)	& -	   &\([512,4,4]\)\\
\midrule
Eq. MaxPool      &\([512,4,4]\)	& $2$	   &\([512,1,1]\)\\
\midrule
Fully Connected     &\([1024]\)	& -	   &\([10]\)\\
  \bottomrule
 \end{tabular}
\end{table}

\end{document}